\documentclass[letterpaper, 10 pt, conference]{ieeeconf}  

\IEEEoverridecommandlockouts                              

\usepackage{graphicx} 
\usepackage{xcolor}
\usepackage{times} 
\usepackage{amsmath} 

\usepackage{array}
\usepackage{blkarray}
\usepackage{multicol}
\usepackage{enumerate}
\usepackage[bookmarks=true]{hyperref}

\newcommand{\bs}[1]{\boldsymbol{#1}}

\newcolumntype{P}[1]{>{\centering\arraybackslash}p{#1}}

\usepackage{geometry}
\title{\LARGE \bf
Search Methods for Policy Decompositions
}

\author{Ashwin Khadke and Hartmut Geyer
\thanks{Ashwin Khadke and Hartmut Geyer are with the Robotics Institute,
        Carnegie Mellon University, 5000 Forbes Avenue, Pittsburgh PA, 15213 USA
        {\tt\small \{akhadke, hgeyer\}@andrew.cmu.edu}}%
}

\begin{document}
\maketitle
\thispagestyle{empty}
\pagestyle{empty}
\begin{abstract}

Computing optimal control policies for complex dynamical systems requires approximation methods to remain computationally tractable. Several approximation methods have been developed to tackle this problem. However, these methods do not reason about the suboptimality induced in the resulting control policies due to these approximations. We introduced Policy Decomposition, an approximation method that provides a suboptimality estimate, in our earlier work. Policy decomposition proposes strategies to break an optimal control problem into lower-dimensional subproblems, whose optimal solutions are combined to build a control policy for the original system. However, the number of possible strategies to decompose a system scale quickly with the complexity of a system, posing a combinatorial challenge. In this work we investigate the use of Genetic Algorithm and Monte-Carlo Tree Search to alleviate this challenge. We identify decompositions for swing-up control of a 4 degree-of-freedom manipulator, balance control of a simplified biped, and hover control of a quadcopter.
\end{abstract}

\section{Introduction}

Computing global optimal control policies for complex systems quickly becomes intractable owing to the curse of dimensionality \cite{Bertsekas:1995}. One approach to tackle this problem is to compute several local controllers using iLQG \cite{Todorov:2005} or DDP \cite{Tassa:2014} and combine them to obtain a global policy \cite{Atkeson:2013, Tedrake:2010, Zhong:2013}. Another approach is to compute policies as functions of some lower-dimensional features of the state; either hand designed \cite{Stilman:2005}, or obtained by minimizing some projection error \cite{Bouvrie:2017}. As can be inferred from Tab.~\ref{tab:relatedworks}, none of these methods predict nor reason about the effect of the respective simplifications on the closed-loop behavior of the resulting policy. For linear systems or linear approximations of nonlinear systems,
\cite{Xue:2016} and \cite{Alla:2017} identify subspaces within the state-space for which the optimal control of the lower-dimensional system and that of the original system closely resemble. 
Reduction methods that assess the suboptimality of the resulting policies when applied to the full nonlinear system have not been explored. 
\begin{table}[h!]
\caption{Comparison of methods to solve the optimal control problem}
\label{tab:relatedworks}
\centering
\footnotesize
\begin{tabular}{P{3.1cm}|P{1.2cm}|P{1.6cm}|P{1.1cm}}
\textbf{method} & \textbf{output} & \textbf{suboptimality} &  \textbf{run time} \\
\hline
iLQG \cite{Todorov:2005} / DDP \cite{Tassa:2014} & Trajectory & unknown & $O(\text{d}^3)$\\
Trajectory based DP\cite{Atkeson:2013} \cite{Tedrake:2010} & Policy & unknown & unknown\\
Model Reduction \cite{Stilman:2005}\cite{Bouvrie:2017} & Policy & unknown & $O(n^{\text{d}_{\text{reduced}}})$ \\
Policy Decomposition \cite{Khadke:2021} & Policy & estimated & $O(n^{\text{d}_{\text{reduced}}})$
\end{tabular}
\end{table}

\begin{figure}
  \centering
  \includegraphics[width=0.46\textwidth]{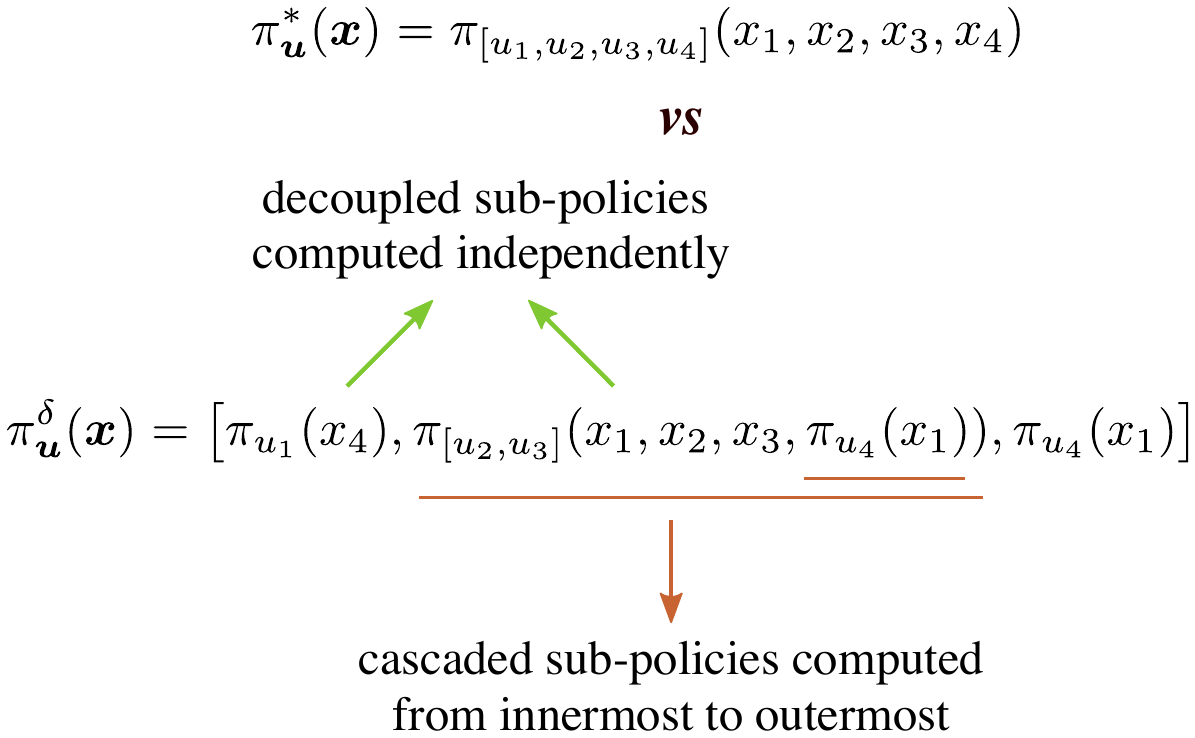}
  \caption{Idea of Policy Decomposition shown for a fictive system. 
  Instead of jointly computing optimal control policies for all inputs over the full state-space (\textbf{top}), the policies are approximated with decoupled and cascaded sub-policies for smaller sub-systems that are faster to compute (\textbf{bottom}). 
  }
  \label{f:decomp_example}
\end{figure}

In our earlier work we introduced Policy Decomposition, an approximate method for solving optimal control problems with suboptimality estimates for the resulting control policies \cite{Khadke:2021}. Policy Decomposition proposes strategies to decouple and cascade the process of computing control policies for different inputs to a system. Based on the strategy the control policies depend on only a subset of the entire state of the system (Fig.~\ref{f:decomp_example}), and can be obtained by solving lower-dimensional optimal control problems leading to reduction in policy compute times. We further introduced the \emph{value error}, i.e. the difference between value functions of control policies obtained with and without decomposing, to assess the closed-loop performance of possible decompositions. This error cannot be obtained without knowing the true optimal control, and we estimate it based on local LQR 
approximations. The estimate can be computed without solving for the policies and in minimal time, allowing us to \emph{predict} which decompositions sacrifice minimally on optimality while offering substantial reduction in computation.

\begin{figure}[h]
    \centering
    \includegraphics[width=0.46\textwidth]{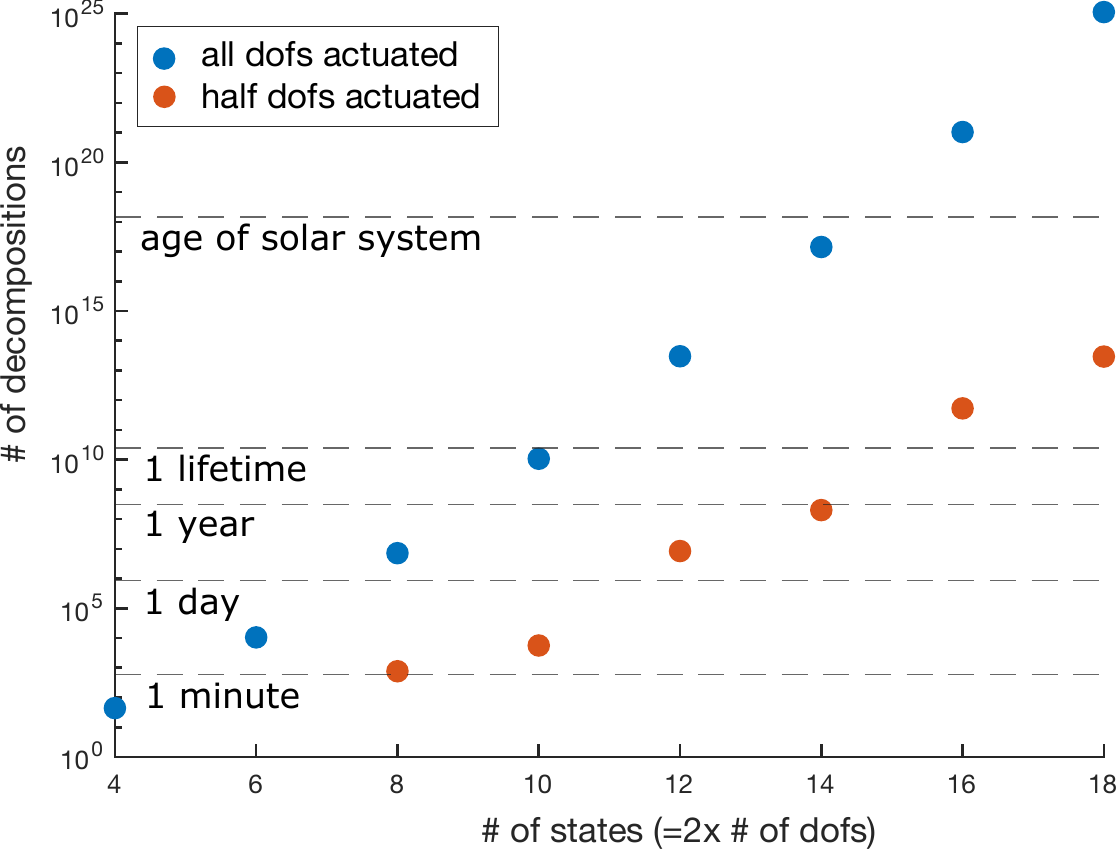}
    \caption{Number of possible decompositions as the number of state variables and control inputs increase. Assuming it takes \textbf{0.1sec/value error estimate} computation, the number of decompositions that can be evaluated in some fixed times are marked with dotted lines.}
    \label{f:decompcount}
\end{figure}

The number of possible decompositions that can be constructed with just the two fundamental strategies of cascading and decoupling shown in Fig.~\ref{f:decomp_example}, turns out to be substantial even for moderately complex systems. Exhaustive search by evaluating the value error estimates for all possible decompositions becomes impractical for systems with more than $12$ state variables (Fig.~\ref{f:decompcount}). We thus explore the use of Genetic Algorithm (GA) \cite{Mitchell:1998} and Monte-Carlo Tree Search (MCTS) \cite{Browne:2012} to efficiently find promising decompositions. We first review the main ideas behind Policy Decomposition in Sec.~\ref{s:policy_decomposition}.  Sec.~\ref{s:search_methods} introduces input-trees, an abstraction to represent any decomposition, and discusses how GA and MCTS search over the space of input-trees. We find decompositions for the control of a 4 degree-of-freedom manipulator, a simplified biped and a quadcopter in Sec.~\ref{s:results}, and conclude with some directions for future work in Sec.~\ref{sec:conclusion}.
\section{Policy Decomposition}
\label{s:policy_decomposition}
Consider the general dynamical system 
\begin{equation} \label{eq:dynamics}
\dot{\bs{x}} = \bs{f}(\bs{x}, \boldsymbol{u})
\end{equation} 
with state $\bs{x}$ and input $\bs{u}$. The optimal control policy $\pi^*_{\bs{u}}(\bs{x})$ for this system minimizes the objective function
\begin{equation} \label{eq:cost}
J_{} = \int_0^\infty e^{-\lambda t} c(\bs{x}(t), \bs{u}(t)) \: dt.
\end{equation}
which describes the discounted sum of costs $c(\bs{x},\bs{u}) = (\bs{x} - \bs{x}^d)^T \bs{Q} (\bs{x} - \bs{x}^d) + (\bs{u}-\bs{u}^d)^T \bs{R} (\bs{u}-\bs{u}^d)$ accrued over time, where $\bs{x}^d$ is the goal state, and $\bs{u}^d$ the input that stabilizes the system at $\bs{x}^d$. The discount factor $\lambda$ characterizes the trade-off between immediate and future costs. 

\begin{figure*}[h!]
  \centering
  \includegraphics[width=\textwidth]{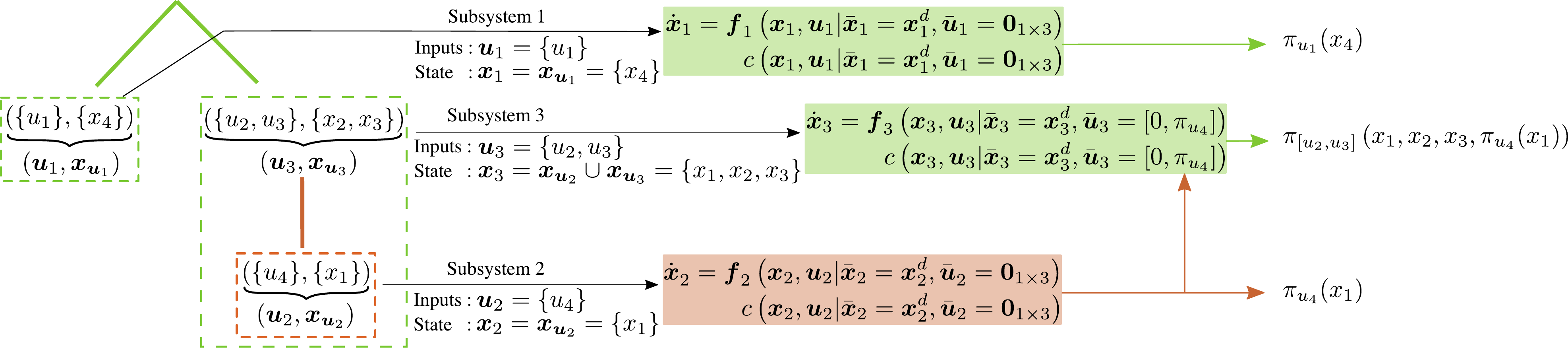}
  \caption{Input-tree (\textbf{left}) for the decomposition shown in Fig~\ref{f:decomp_example} and the resulting subsystems (\textbf{mid}) are depicted. Policies for $u_1$ and $u_4$, i.e. inputs at the leaf nodes, are obtained first by independently solving optimal control problems for subsystems 1 and 2 respectively. Policies for $u_2$ and $u_3$ are computed jointly by solving the optimal control problem for subsystem 3. Resulting policies for different inputs (\textbf{right}) are a function of different state variables.
  }
  \label{f:input_tree}
\end{figure*}
Policy Decomposition approximates the search for one high-dimensional policy $\pi^*_{\bs{u}}(\bs{x})$ with a search for a combination of lower-dimensional sub-policies of smaller optimal control problems that are much faster to compute. These sub-policies are computed in either a cascaded or decoupled fashion and then reassembled into one decomposition policy $\pi^\delta_{\bs{u}}(\bs{x})$ for the entire system (Fig.~\ref{f:decomp_example}). 
Any sub-policy $\pi_{\bs{u}_i}(\bs{x}_i)$ is the optimal control policy for its corresponding subsystem, 
\begin{equation}
\dot{\bs{x}}_i = \bs{f}_i (\bs{x}_i, \bs{u}_i \mid\: \bar{\bs{x}}_i=\bar{\bs{x}}_i^d, \bar{\bs{u}}_i),
\end{equation}
where $\bs{x}_i$ and $\bs{u}_i$ are subsets of $\bs{x}$ and $\bs{u}$, $\bs{f}_i$ only contains the dynamics associated with $\bs{x}_i$, and the complement state $\bar{\bs{x}}_i = \bs{x} \setminus \bs{x}_i$ is assumed to be constant. The complement input $\bar{\bs{u}}_i = \bs{u}\setminus\bs{u}_i$ comprises of inputs that are decoupled from $\bs{u}_i$ and those that are in cascade with $\bs{u}_i$. The decoupled inputs are set to zero and sub-policies for the cascaded inputs are used as is while computing $\pi_{\bs{u}_i}(\bs{x}_i)$. In general, $\pi_{\bar{\bs{u}}_i}(\bs{x}_i) = [0,\: \ldots,\: 0,\: \pi_{\bs{u}_j}(\bs{x}_j), \:0 ,\: \ldots,\: 0]$ where $\bs{u}_j\subseteq\bar{\bs{u}}_i$ are inputs that appear lower in the cascade to $\bs{u}_i$, and $\bs{x}_j\subseteq\bs{x}_i$. Note, (i) $\pi_{\bar{\bs{u}}_i}(\bs{x}_i)$ can contain multiple sub-policies, (ii) these sub-policies can themselves be computed in a cascaded or decoupled fashion, and (iii) they have to be known before computing $\pi_{\bs{u}_i}(\bs{x}_i)$. We represent policies with lookup tables and use Policy Iteration \cite{Bertsekas:1995} to compute them. Dimensions of the tables are suitably reduced when using a decomposition.


Value error, $\text{err}^\delta$, defined as the average difference over the state space $\mathcal{S}$ between the value functions $V^\delta$ and $V^*$ of the control policies obtained with and without decomposition, 
\begin{equation} \label{e:ValueError}
    \text{err}^\delta = \frac{1}{|\mathcal{S}|}\: \int_\mathcal{S} V^\delta(\boldsymbol{x}) - V^*(\boldsymbol{x}) \:\: d\boldsymbol{x},
\end{equation}
directly quantifies the suboptimality of the resulting control. 
$\text{err}^\delta$ cannot be computed without knowing $V^*$, the optimal value function for the original intractable problem. Thus, an estimate is derived by linearizing the dynamics (Eq.~\ref{eq:dynamics}) about the goal state and input, $\dot{\bs{x}} = \bs{A}(\bs{x} - \bs{x}^d) + \bs{B}(\bs{u} - \bs{u}^d)$. As the costs are quadratic, the optimal control of this linear system is an LQR, whose value function $V_{\text{lqr}}^*(\bs{x})$ can be readily computed \cite{Palanisamy:2015}. Value error estimate of decomposition $\delta$ is
\begin{equation} \label{e:LQREstimate}
\text{err}^\delta_\text{lqr} = \frac{1}{|\mathcal{S}|}\: \int_\mathcal{S} V^\delta_\text{lqr}(\boldsymbol{x}) - V^*_\text{lqr}(\boldsymbol{x}) \:\: d\boldsymbol{x},
\end{equation}
where $V_{\text{lqr}}^{\delta}(\bs{x})$ is the value function for the equivalent decomposition of the linear system \cite{Khadke:2021}. Note that, $\text{err}^\delta_{\text{lqr}}\in[0,\infty]$.  

\section{Searching For Promising Decompositions}
\label{s:search_methods}
We first introduce an intuitive abstraction for decompositions that allows us to search over the space of possibilities. A decomposition $\delta$ can be represented using an input-tree $T^\delta=(\mathcal{V},\mathcal{E})$ where all nodes except the root node are tuples of disjoint subsets of inputs and state variables $\bs{v}_i = \left(\bs{u}_{i}, \bs{x}_{\bs{u}_i}\right) \forall \bs{v}_i \in \mathcal{V} \setminus \{\bs{v}_{\text{root}}\}$. Inputs that lie on the same branch are in a cascade where policies for inputs lower in the branch (leaf node being the lowest) influence the policies for inputs higher-up. Inputs that belong to different branches are decoupled for the sake of policy computation. A sub-tree rooted at node $\bs{v}_i$ characterizes a subsystem with control inputs $\bs{u}_i$ and state $\bs{x}_i = \cup_j\bs{x}_{\bs{u}_j}$, where $\bs{x}_{\bs{u}_j}$ are state variables belonging to nodes in the sub-tree. Note, $\bs{x}_{\bs{u}_i} \subseteq \bs{x_i}$ and $\bs{x}_{\bs{u}_i}$ may be empty if it does not belong to a leaf-node. Fig.~\ref{f:input_tree} depicts the input-tree for the decomposition shown in Fig.~\ref{f:decomp_example} and describes the resulting subsystems. Policies are computed in a child-first order, starting from leaf nodes followed by their parents and so on. Finally, input-trees can be easily enumerated, allowing us to count the number of possible decompositions for a system (See Sec.~\ref{apx:decomp_count}).

Search for promising decompositions, requires algorithms that can operate over the combinatorial space of possibilities. Several methods have been proposed to tackle such problems \cite{Glover:1998, Blum:2003}. Genetic Algorithm (GA) is one such method that has been used for solving a variety of combinatorial optimization problems \cite{Chapman:1994, Stanley:2002}. 
GA starts off with a randomly generated set of candidate solutions and suitably combines promising candidates to find better ones. 
A top-down alternative to GA is to start from the original system and decompose it step-by-step. The search for good decompositions can then be posed as a sequential decision making problem for which a number of algorithms exist, especially in the context of computer games \cite{Bouzy:2001}. Among these, Monte-Carlo Tree Search (MCTS) \cite{Browne:2012} methods have been shown to be highly effective for games that have a large number of possible moves in each step, a similar challenge that we encounter when generating decompositions from the original system. Here, we describe the requisite adaptations to GA and MCTS for searching over the space of input-trees.

\subsection{Genetic Algorithm}
\label{ss:ga}

GA iteratively evolves a randomly generated initial population of candidates, through selection, crossovers and mutations, to find promising ones. In this case, the candidates are input-trees. We generate the initial population by uniformly sampling from the set of all possible input-trees for a system (Sec.~\ref{ss:randsamp}). The quality of an input tree is assessed using a fitness function that accounts for value error estimates as well as estimates of the policy compute times for the corresponding decomposition. Moreover, we add a memory component to GA using a hash table, that alleviates the need to recompute fitness values for previously seen candidates. Typical GA components include operators for crossover and mutation. We introduce mutation operators for input-trees but exclude crossovers as we did not observe any improvement in search performance on including them.

\subsubsection{Fitness Function}
The fitness function for a decomposition $\delta$ is a product of two components 
\begin{equation}
\label{eq:fitnessfunc}
    F(\delta) = F_{\text{err}}(\delta)\times F_{\text{comp}}(\delta)
\end{equation}
where $F_{\text{err}}(\delta)$ and $F_{\text{comp}}(\delta)$ quantify the suboptimality of a decomposition and potential reduction in policy computation time respectively. $F_{\text{err}}(\delta) = (1 - \exp(-\text{err}^\delta_{\text{lqr}}))$, is the value error estimate (Eq.~\ref{e:LQREstimate}) scaled to the range $[0,1]$, and $F_{\text{comp}}(\delta)$ is the ratio of estimates of floating point operations required to compute policies with and without decomposition $\delta$. We use lookup tables to represent policies and policy iteration \cite{Bertsekas:1995} to compute them, and $F_{\text{comp}}(\delta)$ can thus be derived using the size of the resulting tables, maximum iterations for policy evaluation and update, and number of actions sampled in every iteration of policy update. Note that $F(\delta)\in[0,1]$ where lower values indicate a more promising decomposition.

\subsubsection{Mutations}
\label{sss:mutation}
\begin{figure*}[h!]
    \centering
    \includegraphics[width=0.8\textwidth]{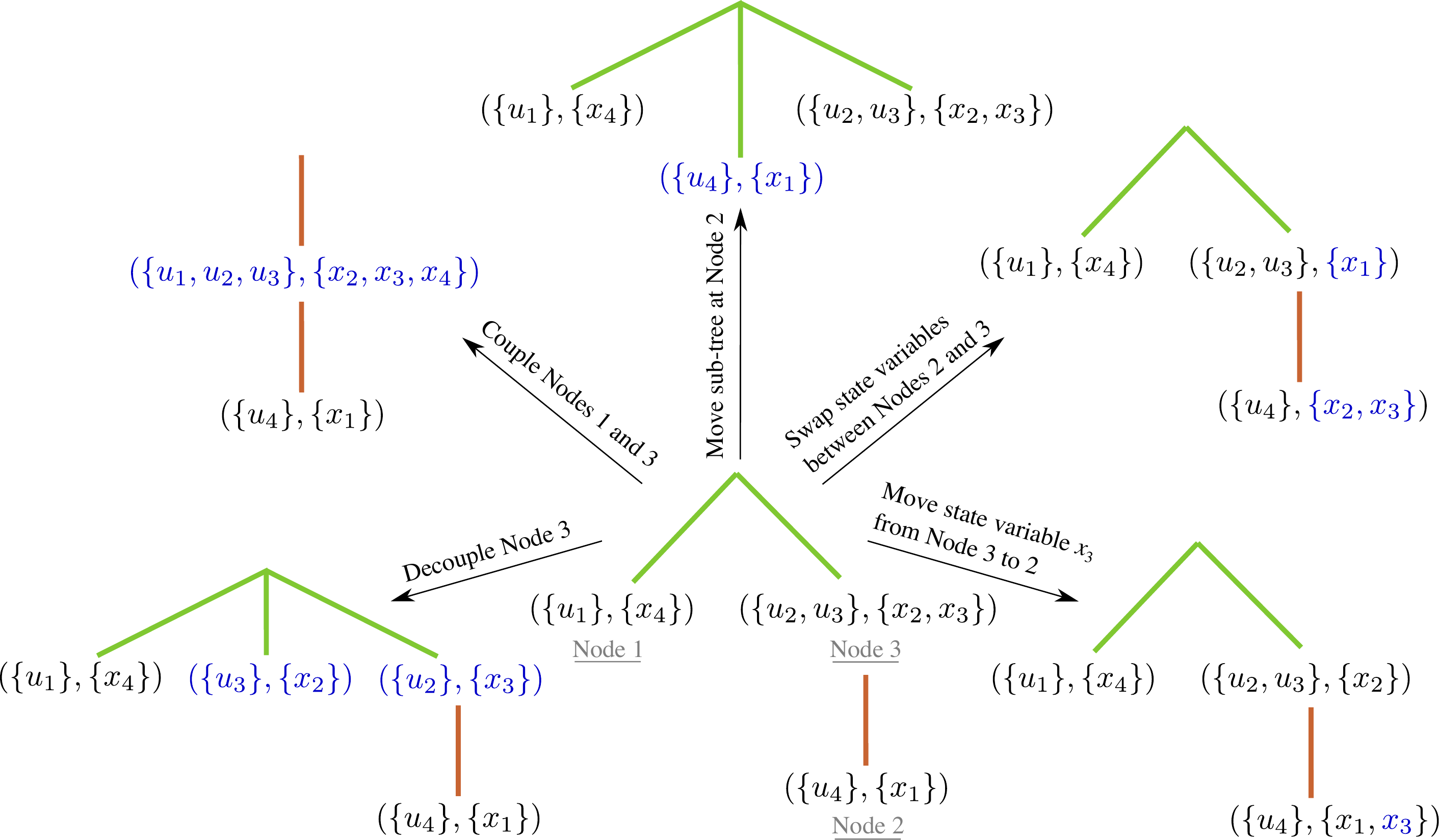}
    \caption{Mutation operators for Genetic Algorithm described in Sec.~\ref{sss:mutation} applied to the input-tree depicted in Fig.~\ref{f:input_tree}.}
    \label{f:mutations}
\end{figure*}

Valid mutations to input-trees are
\begin{enumerate}[(i)]
    \item Swap state variables between nodes
    \item Move a single state variable from one node to another
    \item Move a sub-tree
    \item Couple two nodes
    \item Decouple a node into two nodes
\end{enumerate}
Fig.~\ref{f:mutations} depicts the above operations applied to the input-tree depicted in Fig.~\ref{f:input_tree}. In every GA iteration, a candidate selected for mutation is modified using only one operator. Operators (i), (ii) and (iii) each have a 25$\%$ chance of being applied to a candidate. If neither of these three operators are applied then two distinct inputs are randomly selected and depending on whether they are decoupled or coupled operators (iv) or (v) are applied respectively.

\subsubsection{Hashing Input-Trees}
To avoid re-evaluating the fitness for previously seen candidates, we maintain a hash table \cite{Cormen:2009}. This hash table maps a unique identifier, or a \emph{key}, for input-trees to their computed fitness values. See Sec.~\ref{apx:hashkeys} for details on computing these unique identifiers for input-trees.
\subsection{Monte-Carlo Tree Search}
\label{ss:mcts}

MCTS takes the top-down approach, and creates a search tree of input-trees with an input-tree representing the undecomposed system at its root. It builds the search tree by doing rollouts i.e. a sequence of node expansions starting from the root till a terminal node is reached. After each rollout a backup operation is performed whereby a value function for every node in the explored branch is computed/updated. The value function for a node captures the quality of solutions obtained in the sub-tree rooted at that node, and is used during rollouts to select nodes for expansion. Here, we describe the strategy for expanding nodes, the role of value functions during rollouts, the backup operation performed at the end of a rollout, and some techniques to speed-up search.
\subsubsection{Node Expansion}
\begin{figure*}[t!]
    \centering
    \includegraphics[width=0.9\textwidth]{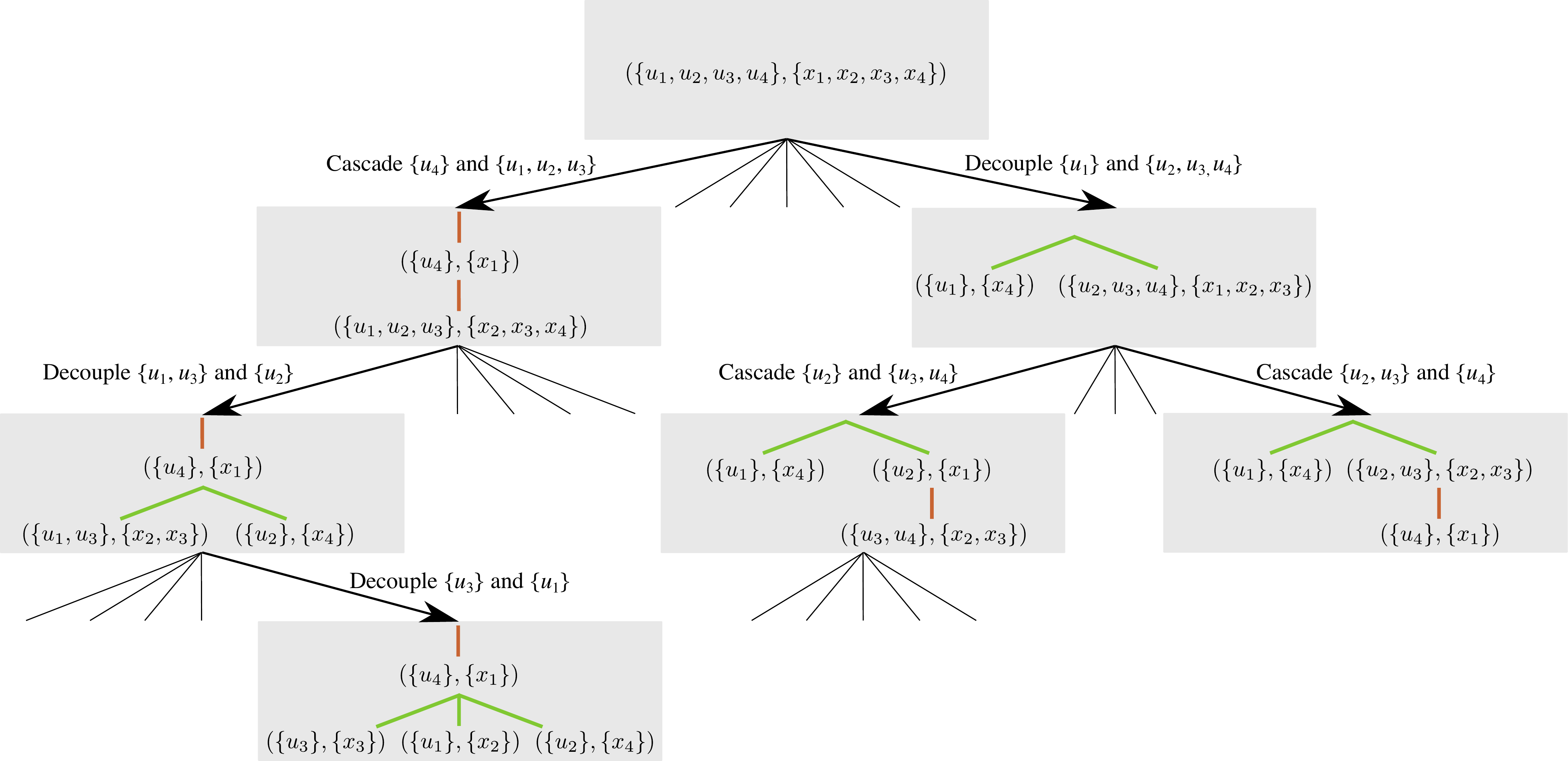}
    \caption{Example of a search tree generated from MCTS rollouts for a fictive system with four inputs and four state variables.}
    \label{f:MCTS_expansion}
\end{figure*}

Expanding a node for decomposition $\delta$ entails evaluating the fitness $F(\delta)$ (Eq.~\ref{eq:fitnessfunc}) and enumerating possible child nodes. These children are constructed by splitting the input and the state variable sets into two, for any of the leafs in $T^\delta$, and arranging them in a cascaded or decoupled fashion. Fig.~\ref{f:MCTS_expansion} depicts an example search tree obtained from MCTS rollouts. We use the UCT strategy \cite{Kocsis:2006},
\begin{equation}\label{e:uct}
    T^{\delta_{\text{expand}}} = \underset{T^{\delta'}\in \text{children}(T^\delta)}{\text{argmin}} Q(T^{\delta'}) - \sqrt{\frac{2 \ln(N_{T^\delta}+1)}{N_{T^{\delta'}}}}
\end{equation}
to pick a node to expand. Here, $Q(T^{\delta'})$, i.e. the value function for the node in the search tree corresponding to $T^{\delta'}$, is the best fitness value observed in the sub-tree rooted at that node. $Q(T^{\delta'})$ is initialized to $F(\delta')$. $N_{T^{\delta'}}$ denotes the number of times $T^{\delta'}$ was visited. We break ties randomly.
\subsubsection{Backup}
At the end of each rollout the value functions for nodes visited during the rollout are updated as follows
\begin{equation}
Q(T^\delta) = \underset{T^{\delta'}\in\left(\{T^\delta\} \hspace{0.1cm}\cup \hspace{0.1cm}\text{children}(T^\delta)\right)}{\text{min}} Q(T^{\delta'})\nonumber
\end{equation}
starting from the terminal node and going towards the root of the search tree. Value function of the root node is the fitness value of the best decomposition identified in the search.
\subsubsection{Speed-up Techniques}
To prevent redundant rollouts, we remove a node from consideration in the UCT strategy (Eq.~\ref{e:uct}) if all possible nodes reachable through it have been expanded. Similar to GA, we use a hash table to avoid re-evaluating fitness for previously seen input-trees.
\subsection{Uniform Sampling of Input-Trees}
\label{ss:randsamp}
The strategy to sample input-trees is tightly linked to the process of constructing them, which entails partitioning the set of control inputs into groups, arranging the groups in a tree, and then assigning the state variables to nodes of the tree. To ensure uniform sampling, probabilities of choosing a partition and tree structure are scaled proportional to the number of valid input-trees resulting from said partition and structure. Finally, an assignment of state variables consistent with the tree structure is uniformly sampled (Sec.~\ref{apx:uniformsampling}).
\section{Results and Discussion}
\label{s:results}
We test the efficacy of GA and MCTS in finding promising policy decompositions for three representative robotic systems, a 4 degree-of-freedom manipulator (Fig.~\ref{f:manip4dof}(a)), a simplified model of a biped (Fig.~\ref{f:biped}(a)), and a quadcopter (Fig.~\ref{f:quadcopter}(a)). We run every search algorithm 5 times for a period of 300, 600 and 1200 seconds for the biped, manipulator and quadcopter respectively. We report the number of unique decompositions discovered. Moreover, we derive policies for the best decomposition, i.e. the one with the lowest fitness value (Eq.~\ref{eq:fitnessfunc}), in each run and report the suboptimality estimates (Eq.~\ref{e:LQREstimate}) as well as the policy computation times (Tab.~\ref{tab:searchstats}). Tabs.~\ref{tab:biped_compute}, \ref{tab:4linkpendula_compute} and \ref{tab:quadcopter_compute} list the best decompositions across all runs for the three systems.


Overall, in comparison to MCTS and random sampling GA consistently identifies decompositions that are on par, if not better in terms of suboptimality estimates, but offer greater reduction in policy computation times, and discovers the most number of unique decompositions. Furthermore, for the biped and the manipulator, GA evaluates only a small fraction of all possible decompositions but consistently identifies the one with the \emph{lowest} fitness value. Here, we analyze the decompositions identified through the search methods. For details regarding policy computation see Sec.~\ref{apx:policycomputation}. 


\subsubsection{\textbf{Biped}}
\begin{figure}
    \centering
    \includegraphics[width=0.49\textwidth]{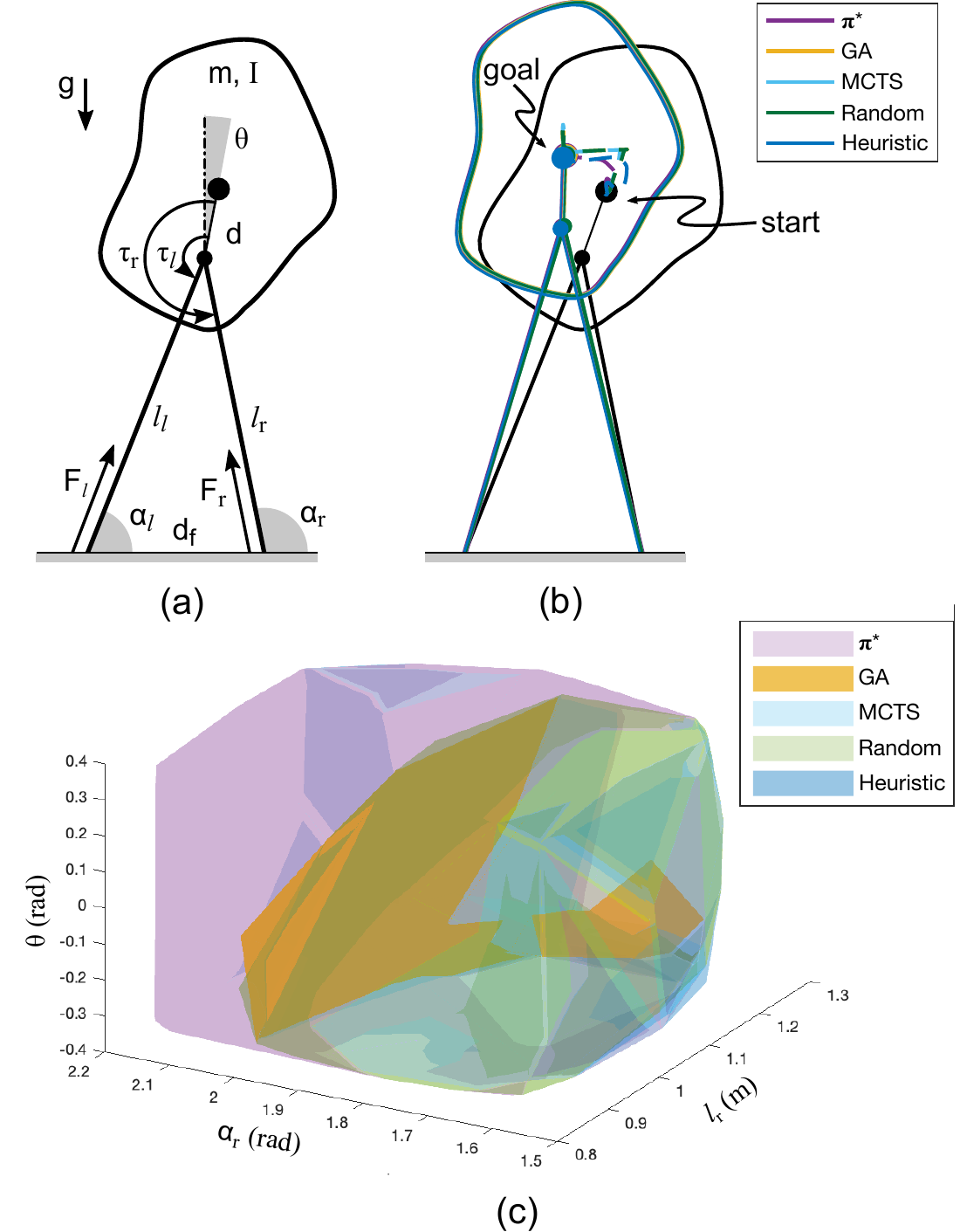}
    \caption{(a) Simplified model of a biped. See Sec.~\ref{apx:biped} for details. (b) Closed-loop trajectories derived from the optimal policy (\textbf{magenta}), heuristic policy (\textbf{blue}), and from policies computed using the best decompositions found by GA (\textbf{yellow}), MCTS (\textbf{cyan}) and random sampling (\textbf{green}). (c) Basins of attraction, estimated by simulating the system with the policies from different decompositions.}
    \label{f:biped}
\end{figure}
We design policies to balance the biped (Fig.~\ref{f:biped}(a)) in double stance. Sec.~\ref{apx:biped} details the biped dynamics. From Tab~\ref{tab:biped_compute}, the best decompositions discovered by GA and MCTS regulate the vertical behavior of the biped using leg force $F_l$, the fore-aft behavior using the other leg force $F_r$ and a hip-torque $(\tau_l/\tau_r)$, and regulate the torso using the remaining hip torque. These decompositions are similar to the heuristic approach (Tab.~\ref{tab:biped_compute}, row 5) of regulating the centroidal behavior using leg forces and computing a torso control in cascade \cite{Martin:2017}, 
but offer major reduction in complexity. It takes about 1.7 hours to derive the optimal policy whereas decompositions found by GA and MCTS reduce the computation time to seconds (Tab.~\ref{tab:searchstats}, row 1). But, the reduction comes at the cost of smaller basins of attraction (Fig.~\ref{f:biped}(c)).  
\begin{table}[h!]
\caption{Best decompositions found by search methods : \textbf{biped}}
\label{tab:biped_compute}
\centering
\footnotesize
\begin{tabular}{P{0.9cm}|P{5.3cm}|P{1.4cm}}
\textbf{method} & $\bs{\delta}$ \textbf{with the lowest fitness value} & $\bs{\#}$\textbf{policy parameters} \\
\hline
& \scriptsize{$\pi^*$}& 52,865,904\\
GA & \scriptsize{$[\pi_{\tau_r}(\theta, \dot{\theta}), \pi_{\tau_l}(l_r), \pi_{\text{F}_r}(l_r, \dot{z}, \pi_{\tau_l}), \pi_{\text{F}_l}(\alpha_r, \dot{x})]$} & 736 \\
MCTS & \scriptsize{$[\pi_{\tau_r}(l_r), \pi_{\tau_l}(\theta, \dot{\theta}), \pi_{\text{F}_r}(l_r, \dot{z}, \pi_{\tau_r}), \pi_{\text{F}_l}(\alpha_r, \dot{x})]$} & 736\\
Random & \scriptsize{$[\pi_{\tau_r}(\theta, \dot{\theta}), \pi_{\tau_l}(\alpha_r), \pi_{\text{F}_r}(l_r, \dot{z}), \pi_{\text{F}_l}(\alpha_r, \dot{x}, \pi_{\tau_l})]$} & 736\\
Heuristic & \scriptsize{$[\pi_{[F_l,F_r]}(l_r,\alpha_r,\dot{x}, \dot{z})$, $\pi_{[\tau_l,\tau_r]}(l_r,\alpha_r,\dot{x}, \dot{z}, \theta, \dot{\theta}, \pi_{[F_l, F_r]})]$} & 26,522,860\\
\end{tabular}
\end{table}
\subsubsection{\textbf{Manipulator}}
\begin{figure}[h!]
\centering
    \includegraphics[width=0.49\textwidth]{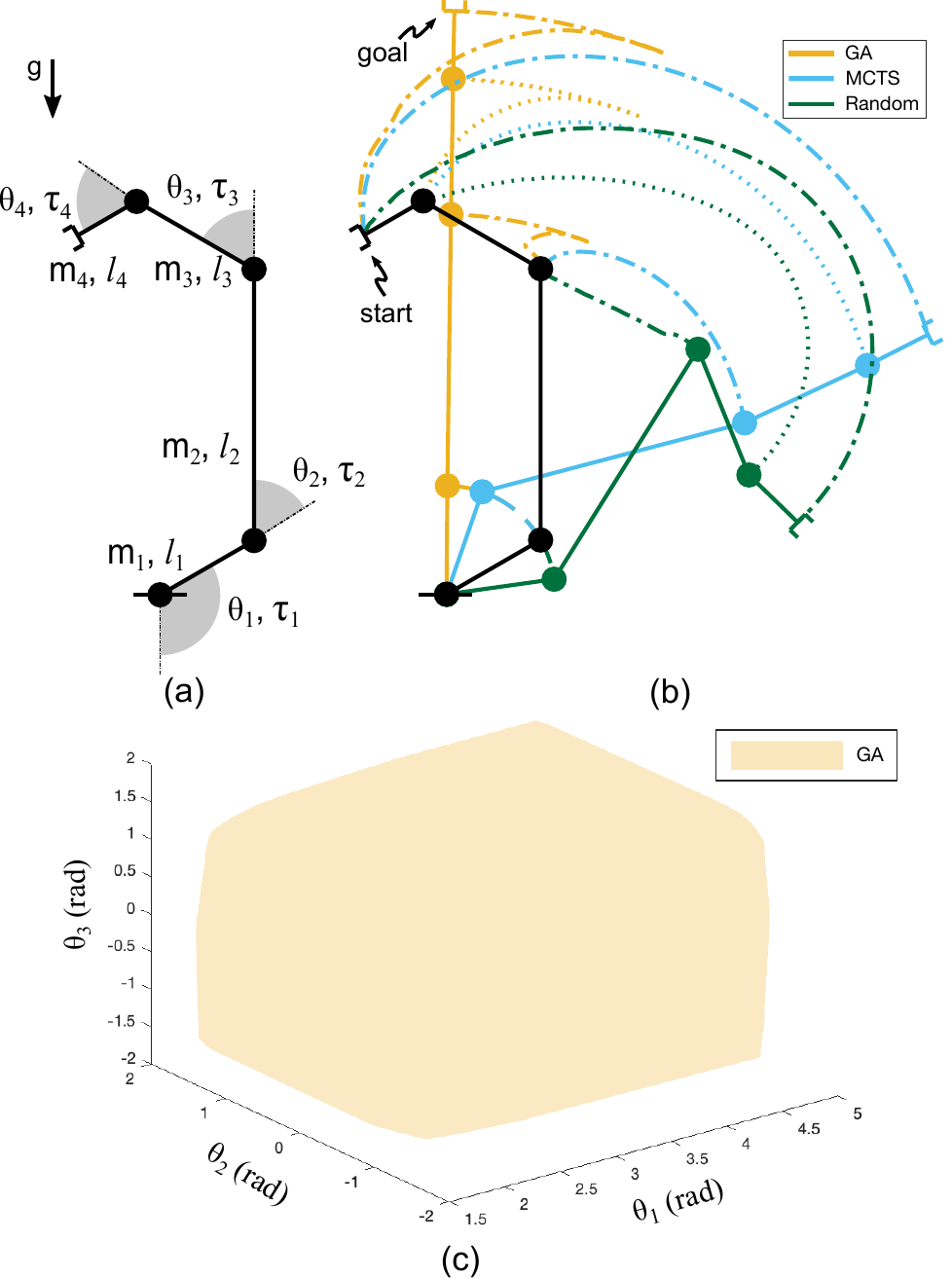}
    \caption{(a) Model of a 4 degree-of-freedom manipulator. See Sec.~\ref{apx:manip} for details. (b) Closed-loop trajectories for swing-up control; derived from policies obtained using the best decompositions found by GA (\textbf{yellow}), MCTS (\textbf{cyan}) and random sampling (\textbf{green}) are shown. (c) Estimated basin of attraction for policies from decomposition found by GA. }
    \label{f:manip4dof}
\end{figure}
For the manipulator (Fig.~\ref{f:manip4dof}(a)), we desire control policies that drive it to an upright position. Sec.~\ref{apx:manip} details the dynamics parameters. Due to computational limits, we are unable to derive the optimal policy. 
But, the completely decentralized strategy identified by GA (Tab.~\ref{tab:4linkpendula_compute}, row 2), requires less than 15 seconds to compute torque policies (Tab.~\ref{tab:searchstats}, row 2, column 2) and has a wide basin of attraction (Fig.~\ref{f:manip4dof}(c)). Random sampling too identifies a decentralized strategy (Tab.~\ref{tab:4linkpendula_compute}, row 4), allowing policies to be computed in minimal time, but has a very high value error estimate, resulting in poor performance on the full system (Fig.~\ref{f:manip4dof}(b), green trajectory diverges). The strategy discovered by MCTS couples policy computation for the $1^{\text{st}}$-$2^\text{nd}$ and the $3^{\text{rd}}$-$4^{\text{th}}$ joint torques (Tab.~\ref{tab:4linkpendula_compute}, row 3), and in fact has a lower value error estimate than the one found by GA. But, the resulting policies perform poorly on the full system (Fig.~\ref{f:manip4dof}(b), cyan trajectory diverges). The LQR value error estimate (Eq.~\ref{e:LQREstimate}) only accounts for the linearized system dynamics and is agnostic to the bounds on control inputs, and thus decompositions with low suboptimality estimates need not perform well on the full system. This issue can be alleviated by filtering decompositions shortlisted through search using the DDP based value error estimate discussed in \cite{Khadke:2021}. While the DDP based estimate is better, it too does not provide guarantees but rather serves as an improved heuristic. 

\begin{table}[h!]
\caption{Best decompositions found by search methods : \textbf{manipulator}}
\label{tab:4linkpendula_compute}
\centering
\footnotesize
\begin{tabular}{P{0.8cm}|P{5.6cm}|P{1.38cm}}
\textbf{method} & $\bs{\delta}$ \textbf{with the lowest fitness value} & $\bs{\#}$\textbf{policy parameters} \\
\hline
& \scriptsize{$\pi^*$}& $2.38\times 10^{9}$\\
GA & \scriptsize{$[\pi_{\tau_1}(\theta_1, \dot{\theta}_1), \pi_{\tau_2}(\theta_2, \dot{\theta}_2), \pi_{\tau_3}(\theta_3, \dot{\theta}_3), \pi_{\tau_4}(\theta_4, \dot{\theta}_4)]$} & 884 \\
MCTS & \scriptsize{$[\pi_{[\tau_1, \tau_2]}(\theta_1, \dot{\theta}_1, \theta_2, \dot{\theta}_2), \pi_{[\tau_3, \tau_4]}(\theta_3, \dot{\theta}_3, \theta_4, \dot{\theta}_4)]$} & 97682\\
Random & \scriptsize{$[\pi_{\tau_1}(\theta_1, \dot{\theta}_3), \pi_{\tau_2}(\theta_4, \dot{\theta}_2), \pi_{\tau_3}(\theta_2, \dot{\theta}_4), \pi_{\tau_4}(\theta_3, \dot{\theta}_1)]$} & 884\\
\end{tabular}
\end{table}
\subsubsection{\textbf{Quadcopter}}
\begin{figure}[h!]
    \centering
    \includegraphics[width=0.49\textwidth]{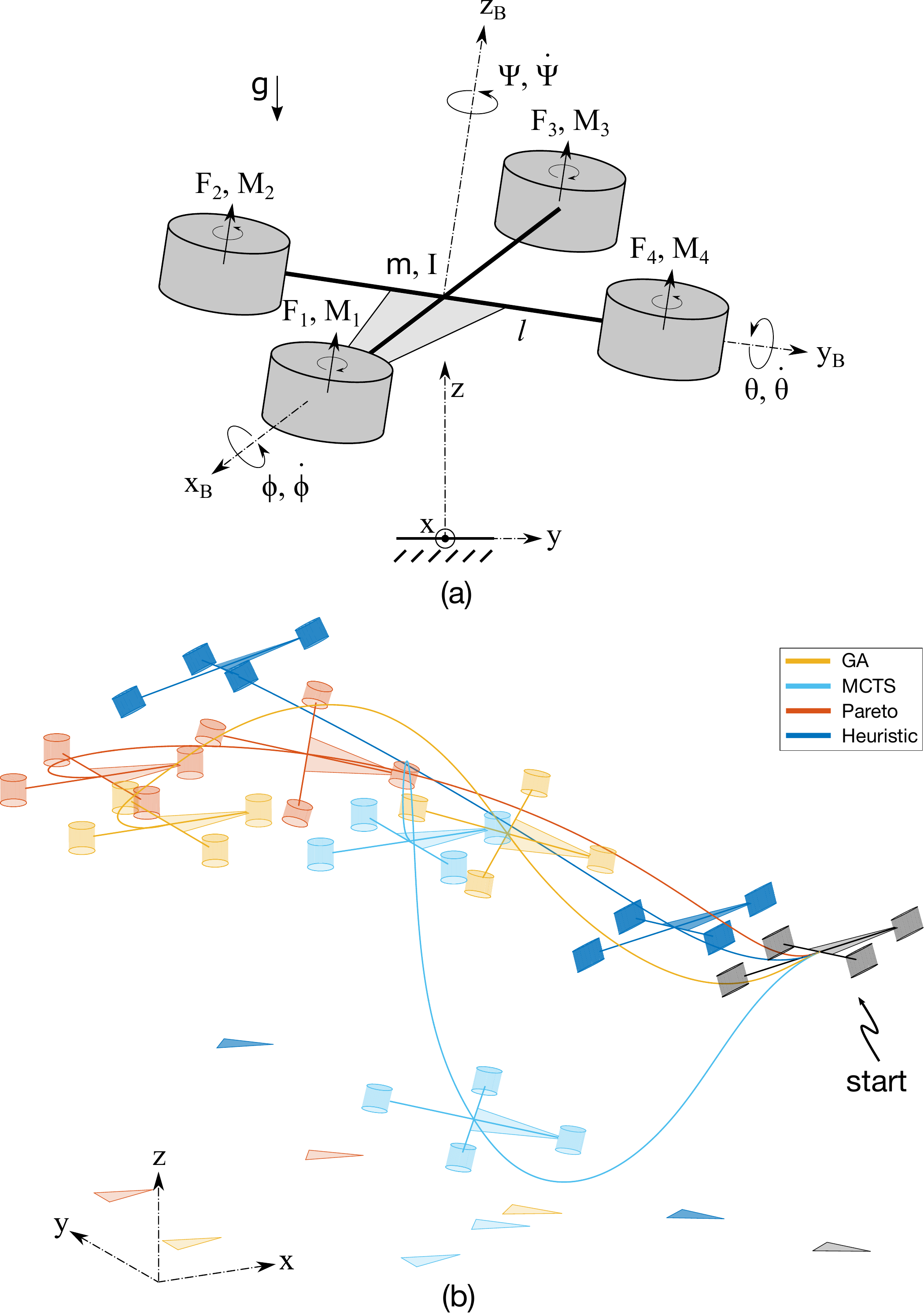}
    \caption{(a) Model of a quadcopter. See Sec.~\ref{apx:quadcopter} for details. (b) Closed-loop trajectories for hover control derived from heuristic policy (\textbf{blue}), and policies computed using best decompositions found by GA (\textbf{yellow}), MCTS (\textbf{cyan}) and from the Pareto front (\textbf{red}) are shown.}
    \label{f:quadcopter}
\end{figure}
The quadcopter (Fig.~\ref{f:quadcopter}(a)) is required to stabilize its attitude and come to rest. We identify decompositions to simplify computing policies for the net thrust $T$ and the roll, pitch and yaw inducing differential thrusts $F_{\phi}$, $F_{\theta}$ and $F_{\psi}$ respectively. Sec.~\ref{apx:quadcopter} details the system modelling. Solving for the optimal policy is beyond our computational limits. However, we discovered almost optimal decompositions with just 20 minutes of search and derived policies using them in less than 3 hours (Tab~\ref{tab:searchstats}, row 3). As depicted in Tab.~\ref{tab:quadcopter_compute}, the best decompositions discovered by GA and MCTS both decouple the yaw $(\psi)$ control. But the one found by GA further cascades $z$ and roll $(\phi)$ control with the pitch control $(\theta)$, offering additional reduction in computation.
Random sampling finds decompositions that have lower suboptimality estimates than the ones found by GA and MCTS (Tab.~\ref{tab:searchstats}, row 3) and lower fitness values as well, but they require significantly higher compute than we can afford. We are thus unable to report policy compute times for these decompositions in Tab.~\ref{tab:searchstats} and also omit it from Tab.~\ref{tab:quadcopter_compute}. In general, our fitness criteria (Eq.~\ref{eq:fitnessfunc}) can skew the search towards decompositions that are almost optimal but offer little/no reduction in computation, or towards those that greatly reduce computation but are highly suboptimal. A workaround is to compute a Pareto front of decompositions over the objectives $F_{\text{comp}}$ and $F_{\text{err}}$ (Eq.~\ref{eq:fitnessfunc}). We derived such a Pareto front using the NSGA II \cite{Deb:2011}, sifted through the front in increasing order of $F_{\text{err}}$, and found a decomposition (Tab.~\ref{tab:quadcopter_compute}, row 4) that fit our computational budget. Closed-loop trajectories for decompositions in Tab~\ref{tab:quadcopter_compute} are shown in Fig.~\ref{f:quadcopter}(b). Decompositions discovered by search are non-trivial and in fact the trivial heuristic (Tab.~\ref{tab:quadcopter_compute}, row 5) to independently regulate roll, pitch, yaw and height with $F_{\phi}$, $F_{\theta}$, $F_{\psi}$ and $T$ respectively, fails to stabilize the attitude and bring the quadcopter to rest (Fig.~\ref{f:quadcopter}, blue).

\begin{table}[h!]
\caption{Best decompositions found by search methods : \textbf{Quadcopter}}
\label{tab:quadcopter_compute}
\centering
\footnotesize
\begin{tabular}{P{0.85cm}|P{5.4cm}|P{1.375cm}}
\textbf{method} & $\bs{\delta}$ \textbf{with the lowest fitness value} & $\bs{\#}$\textbf{policy parameters} \\
\hline
& \scriptsize{$\pi^*$}& $1.74\times 10^{10}$\\
GA & \scriptsize{$[\pi_{F_{\psi}}(\psi, \dot{\psi})$, $\pi_{[T, F_{\phi}]}(z, \dot{z}, \phi, \dot{\phi}, \dot{y}, \dot{\theta})$, $ \pi_{F_\theta}(z, \dot{z}, \phi, \dot{\phi}, \dot{y}, \theta, \dot{\theta}, \dot{x}, \pi_{[T, F_{\phi}]})]$} & 14817796 \\
MCTS & \scriptsize{$[\pi_{F_{\psi}}(\psi, \dot{\psi})$, $\pi_{[T, F_{\phi}, F_{\theta}]}(z, \dot{z}, \phi, \dot{\phi}, \dot{y}, \theta, \dot{\theta}, \dot{x})]$} & 42707812\\
Pareto & \scriptsize{$[\pi_{F_{\psi}}(\psi, \dot{\psi})$, $\pi_{T}(z, \dot{z})$, $\pi_{F_\phi}(z, \dot{z}, \phi, \dot{\phi}, \dot{y}, \pi_{T})$, $\pi_{F_\theta}(z, \dot{z}, \phi, \dot{\phi}, \dot{y}, \theta, \dot{\theta}, \dot{x}, \pi_{T}, \pi_{F_{\phi}})]$} & 14263214\\
Heuristic & \scriptsize{$[\pi_{F_{\psi}}(\psi, \dot{\psi}), \pi_{T}(z, \dot{z}), \pi_{F_\phi}(\phi, \dot{\phi}, \dot{y}), \pi_{F_\theta}(\theta, \dot{\theta}, \dot{x})]$} & 2352\\
\end{tabular}
\end{table}

\begin{table*}
\caption{Summary of search results. $\text{err}^\delta_{\text{lqr}}$ and policy compute times for the best decomposition found (in terms of fitness value Eq.~\ref{eq:fitnessfunc}) as well as the number of unique decompositions discovered are reported, averaged across 5 runs. Algorithms have a fixed time period in each run; 300, 600 and 1200 seconds for the Biped, Manipulator and Quadcopter respectively.}
 \label{tab:searchstats}
\begin{tabular}{P{1.2cm}|P{1.15cm}P{1.165cm}P{1.65cm}|P{1.55cm}P{1.165cm}P{1.5cm}|P{1.55cm}P{1.23cm}P{1.5cm}}
&  & \textbf{GA} & & & \textbf{MCTS} & &  & \textbf{Random} & \\
 & $\textbf{err}_{\text{lqr}}^{\bs{\delta}}$ &  \textbf{time} (sec) & $\bs{\#\delta}$ \textbf{found} & $\textbf{err}_{\text{lqr}}^{\bs{\delta}}$ &  \textbf{time} (sec) & $\bs{\#\delta}$ \textbf{found} & $\textbf{err}_{\text{lqr}}^{\bs{\delta}}$ &  \textbf{time} (sec) & $\bs{\#\delta}$ \textbf{found} \\\hline

\footnotesize{Biped} & \scriptsize{$0.1018\pm 0$} & \scriptsize{$7.35\pm 0$} & \scriptsize{$33785 \pm 591$} & \scriptsize{$0.105\pm0.008$} & \scriptsize{$23 \pm 20.7$} & \scriptsize{$21588 \pm 448$} & \scriptsize{$0.217\pm 0.163$} & \scriptsize{$23 \pm 20.9$} & \scriptsize{$29625 \pm 409$}\\ \hline
\footnotesize{Manipulator} & \scriptsize{$0.0357 \pm 0$} & \scriptsize{$14.94\pm 0$} & \scriptsize{$60135 \pm 2604$} & \scriptsize{$0.087\pm0.167$} & \scriptsize{$821\pm 152$} & \scriptsize{$19993 \pm 297$} & \scriptsize{$16.89\pm22.27$} & \scriptsize{$59.5\pm36.6$} & \scriptsize{$40295 \pm 545$}\\ \hline

\footnotesize{Quadcopter} & \scriptsize{$(1.43 \pm 0)$ $\times10^{-16}$} & \scriptsize{$9534.5\pm 0$} & \scriptsize{$169408 \pm 1475$} & \scriptsize{$(1.08 \pm 0.53)$ $\times10^{-16}$} & - & \scriptsize{$38306 \pm 2832$} & \scriptsize{$(5.72\pm3.36)$ $\times10^{-18}$} & - & \scriptsize{$75176 \pm 1095$}\\ \hline

\end{tabular}
\end{table*}


\section{Conclusions and Future Work}
\label{sec:conclusion}
We addressed the combinatorial challenge of applying Policy Decomposition to complex systems, finding non-trivial decompositions to compute policies for otherwise intractable optimal control problems. We introduced an intuitive abstraction, viz input-trees, to represent decompositions and a metric to gauge their fitness. Moreover, we put forth a strategy to uniformly sample input-trees and operators to mutate them, facilitating the use of GA in search for promising ones. We also offered another perspective to the search problem whereby decompositions are constructed by incrementally decomposing the original complex system, reducing the search to a sequential decision making problem. We showcased how MCTS can be used to tackle it. Across three example systems, viz a simplified biped, a manipulator and a quadcopter, GA consistently finds the best decompositions in a short period of time. Our work makes Policy Decomposition a viable approach to systematically simplify policy synthesis for complex optimal control problems.

We identified three future research directions that would further broaden the applicability of Policy Decomposition. First, to extend the framework to handle optimal trajectory tracking problems. Second, to weigh in the effect of sensing and actuation noise in evaluating the fitness for different decompositions. Finally, the conduciveness to decomposition often depends on the choice of inputs. In case of the quadcopter, policy computation for the rotor forces $F_i$ is not readily decomposable but it is so for the linearly transformed inputs $T$, $F_{\phi}$, $F_{\theta}$ and $F_{\psi}$. Simultaneously discovering mappings from the original inputs to a space favorable for decomposition would be very useful.

\section{APPENDIX}
\subsection{Counting Possible Policy Decompositions}
\label{apx:decomp_count}
For a system with $n$ states and $m$ inputs, the number of possible decompositions resolves to
\begin{equation} \label{e:DecompCount}
\begin{split}
  N(n,m) = \sum_{r=2}^{m} \Bigg\{ \frac{\Delta^m_r}{r!} \sum_{k=1}^r \Bigg[ &\binom{r}{k}\left(\Delta^{r-1}_{r-k} + \Delta^{r-1}_{r-k+1}\right)\\& \sum_{i=0}^{n-k} \binom{n}{i}\Delta^{n-i}_{k}(r-k)^i \Bigg] \Bigg\},
\end{split}
\end{equation}
where $\Delta^a_b = \sum_{c=0}^{b-1} (-1)^c \binom{b}{c} (b-c)^a$. This formula results from the observations that any viable decomposition splits the $m$ inputs of a system into $r$ input groups, where $r$ can range from 2 to $m$, and there are $\Delta^m_r/r!$ ways to distribute the $m$ inputs into $r$ groups \cite{Stanley:1986}; the $r$ input groups can then be arranged into input-trees with $k \in\{1,\dots, r\}$ leaf-nodes leading to $\binom{r}{k}\left(\Delta^{r-1}_{r-k} + \Delta^{r-1}_{r-k+1}\right)$ possible trees per grouping; the $n$ states can be assigned to each such tree with $k$ leaf-nodes in $\sum_{i=0}^{n-k} \binom{n}{i}\Delta^{n-i}_{k}(r-k)^i$ different ways.
\subsection{Hash Keys For Input-Trees}
\label{apx:hashkeys}
For a system with $m$ inputs, we use a binary connectivity matrix $(C\equiv m \times m)$ to encode the graph. 
If input $u_j$ belongs to node $\bs{v}_j$ in the input-tree, then the $j^{\text{th}}$ row in $C$ has entries $1$ for all \emph{other} inputs that belong to $\bs{v}_j$ as well as for inputs that belong to the parent node of $\bs{v}_j$. For the input-tree in Fig.~\ref{f:input_tree} the connectivity matrix is
\begin{equation}
    C = 
\begin{blockarray}{ccccc}
& u_1 & u_2 & u_3 & u_4 \\
\begin{block}{c[cccc]}
 u_1 & 0 & 0 & 0 & 0 \\
 u_2 & 0 & 0 & 1 & 0 \\
 u_3 & 0 & 1 & 0 & 0 \\
 u_4 & 0 & 1 & 1 & 0 \\
\end{block}
\end{blockarray}\nonumber
\end{equation}
We define the binary state-dependence matrix $(S\equiv m\times n)$ to encode the influence of the $n$ state variables on the $m$ inputs. The $j^{\text{th}}$ row in $S$ corresponds to input $u_j$ and has entries $1$ for all state variables that belong to the node $\bs{v}_j$. For the input-tree in Fig.~\ref{f:input_tree} the state-dependence matrix is 

\begin{equation}
    S = 
    \begin{blockarray}{ccccc}
& x_1 & x_2 & x_3 & x_4 \\
\begin{block}{c[cccc]}
 u_1 & 0 & 0 & 0 & 1 \\
 u_2 & 0 & 1 & 1 & 0 \\
 u_3 & 0 & 1 & 1 & 0 \\
 u_4 & 1 & 0 & 0 & 0 \\
\end{block}
\end{blockarray}\nonumber
\end{equation}
The matrices $C$ and $S$ uniquely encode an input-tree.
\subsection{Uniform Random Sampling of Input-Trees}
\label{apx:uniformsampling}
For a system with $m$ inputs and $n$ state variables we first sample $r\in\{2,\cdots,m\}$, i.e. the number of input groups in an input-tree, with probability proportional to the number of possible input-trees with $r$ input groups (each entry in the outermost summation in Eq.~\ref{e:DecompCount}). Next, we generate all partitions of the inputs into $r$ groups 
and pick one uniformly at random. Input groups constitute the sets $\bs{u}_i$ in an input-tree. Subsequently, we sample the number of leaf-nodes $k\in\{1,\cdots,r\}$, with probabilities proportional to the number of input-trees with $r$ input groups and $k$ leaf-nodes (each entry in the second summation in Eq.~\ref{e:DecompCount}). To sample the tree structure we use Prufer Codes \cite{Biggs:1986}. Specifically, we uniformly sample a sequence of numbers of length $r-1$ with exactly $r-k$ distinct entries from $\{1,\cdots,r+1\}$.
Finally, for assigning state variables to different nodes of the input-tree we uniformly sample a label for every variable from $\{1,\cdots,r\}$. Variables with the label $i$ are grouped to form sets $\bs{x}_{\bs{u}_i}$. We re-sample labels if a state assignment is invalid i.e. no variables are assigned to leaf-nodes of the input-tree.

\subsection{Policy Computation Details}
\label{apx:policycomputation}
Here, we report the dynamics parameters for each system and other relevant details. Policies are represented using lookup tables and computed using policy iteration \cite{Bertsekas:1995}. Dimensions of the lookup tables are suitably reduced when using a decomposition versus jointly optimizing the policies. Policies are computed using Matlab on an RTX 2080 Ti. Our code is available at \href{https://github.com/ash22194/Policy-Decomposition}{\texttt{https://github.com/ash22194/ Policy-Decomposition}}
\subsubsection{\textbf{Balancing control for Biped}} \label{apx:biped}
The biped (Fig.~\ref{f:biped}(a)) weighs $m = 72\text{kg}$, has rotational inertia $I=3\text{kgm}^2$, and hip-to-COM distance $d=0.2\text{m}$. Legs are massless and contact the ground at fixed locations $d_f=0.5$m apart. A leg breaks contact if its length exceeds $l_0=1.15$m. In contact,  legs can exert forces ($0\leq F_{l/r}\leq 3mg$) and hip torques ($|\tau_{l/r}|\leq 0.25mg/l_0$) leading to dynamics $m\ddot{x} = F_{r}\cos{\alpha_r} + \frac{\tau_{r}}{l_r}\sin{\alpha_r} + F_{l}\cos{\alpha_l} + \frac{\tau_{l}}{l_l}\sin{\alpha_l}$, $m\ddot{z} = F_{r}\sin{\alpha_r} - \frac{\tau_{r}}{l_r}\cos{\alpha_r} + F_{l}\sin{\alpha_l} - \frac{\tau_{l}}{l_l}\cos{\alpha_l} - mg$, and $I\ddot{\theta} = \tau_{r}(1 + \frac{d}{l_r}\sin(\alpha_r - \theta)) + F_{r}d\cos(\alpha_r - \theta) + \tau_{l}(1 + \frac{d}{l_l}\sin(\alpha_l - \theta)) + F_{l}d\cos(\alpha_l - \theta)$, where $l_l = \sqrt{l_r^2 + d_f^2 + 2l_rd_f\cos{\alpha_r}}$ and $\alpha_l = \arcsin \frac{l_r\sin{\alpha_r}}{l_l}$. The control objective is to balance the biped midway between the footholds.


\subsubsection{\textbf{Swing Up control for Manipulator}} \label{apx:manip} 
We design policies to swing up a fully actuated 4 link manipulator. 
Dynamic parameters are $[m_1,m_2, m_3, m_4] = [5.4, 1.8, 0.6, 0.2]$kg and $[l_1,l_2,l_3,l_4] = [0.2,0.5,0.25,0.125]$m. Torque limits are $|\tau_1|\leq 24$, $|\tau_2|\leq 15$, $|\tau_3|\leq 7.5$ and $|\tau_4|\leq 1$. 

\subsubsection{\textbf{Hover control for a Quadcopter}}
\label{apx:quadcopter} We derived policies to stabilize the attitude of the quadcopter while bringing it to rest at a height of 1m. The system is described by its height $z$, attitude (roll $\phi$, pitch $\theta$ and yaw $\psi$), translational velocities $(\dot{x}, \dot{y}, \dot{z})$, and attitude rate ($\dot{\phi}$, $\dot{\theta}$, $\dot{\psi}$). The quadcopter need not stop at a fixed $xy$ location. 
Inputs to the system are the net thrust $T=\sum_iF_i$ and the roll, pitch and yaw inducing differential thrusts $F_{\phi} = F_4 - F_2$, $F_{\theta} = F_3 - F_1$ and $F_{\psi} = F_2 + F_4 - F_1 - F_3$, with limits $0\leq T\leq 2mg$, $|F_{\phi/\theta}| \leq 0.25mg$ and $F_{\psi} \leq 0.125mg$. Policies for rotor forces $F_i$ can be derived from policies for $T$, $F_{\phi}$, $F_{\theta}$ and $F_{\psi}$ through a linear transformation. The quadcopter weighs 0.5 kg, has rotational inertia $I=\text{diag}([4.86\mathrm{e}{-3}, 4.86\mathrm{e}{-3}, 8.8\mathrm{e}{-3}])$ kgm$^2$, and wing length $l=0.225 \text{m}$. Rotor moments $M_i = k_MF_i$, where $k_M = 0.0383$. Dynamics are described in \cite{Luukkonen:2011}. 
\bibliographystyle{IEEEtran}
\bibliography{references}

\end{document}